\documentclass[a4paper,fleqn]{cas-dc}
\usepackage[numbers]{natbib}
\usepackage{amsmath,amssymb}
\usepackage{booktabs}
\usepackage{multirow}
\usepackage{xcolor}
\usepackage{hyperref}
\usepackage{tikz}
\usepackage[percent]{overpic}
\usepackage{adjustbox}
\usepackage{caption}
\usetikzlibrary{positioning,arrows.meta,shapes.geometric,calc,fit,backgrounds}
\usepackage{graphicx}
\usepackage{relsize}
\begin{document}

\let\WriteBookmarks\relax
\def\floatpagepagefraction{1}
\def\textpagefraction{.001}

\shorttitle{DeepMine-Mamba for Document Image Binarization}
\shortauthors{S.W. Chan}

\title[mode=title]{DeepMine-Mamba: Mitigating Information Dilution in Mamba-Based State Space Models for Document Image Binarization}

\author[1]{Sheng-Wei Chan}[orcid=0009-0002-3983-5163]
\ead{412440330@o365.tku.edu.tw}
\author[1]{Yung-Che Wang}
\ead{614450012@o365.tku.edu.tw}
\author[1]{Hsin-Jui Pan}
\ead{412440314@o365.tku.edu.tw}
\author[1]{Chia-Min Lin}
\ead{170587@o365.tku.edu.tw}
\author[1]{Jen-Shiun Chiang\corref{cor1}}
\ead{chiang@mail.tku.edu.tw}

\affiliation[1]{
  organization={Department of Electrical and Computer Engineering, Tamkang University},
  addressline={No.151, Yingzhuan Rd.},
  city={Tamsui Dist., New Taipei City},
  postcode={251301},
  country={Taiwan}
}

\begin{abstract}
Document image binarization must preserve thin, broken, and low-contrast strokes against degraded backgrounds; Mamba-based state space models have so far been explored only preliminarily for it. We measure what state-space propagation does to document features and find that it does not attenuate foreground responses: every region gains amplitude, but the background gain is $1.8$ times the stroke gain, compressing the foreground-background amplitude contrast from $1.86$ to $1.03$. This compression scales monotonically with the effective receptive field, suggesting it is not specific to state space models. Guided by this, we propose DeepMine-Mamba, whose Anti-Dilution Gate computes its modulation from the propagation-induced feature difference and confines it to stroke regions by a learned stroke prior. Under a strict leave-one-year-out protocol on DIBCO/H-DIBCO with no external training data it is competitive with recent transformer- and gated-convolution baselines; because published protocols differ and are often unstated, we report protocol-controlled evidence separately from cross-paper comparison. Ablations under a fixed objective show that the gate recovers stroke-background discriminability and improves all four metrics, including the two absent from the objective.
\end{abstract}


\maketitle

\section{Introduction}
Document image binarization converts degraded document images into binary foreground-background maps for optical character recognition, historical manuscript restoration, and document image analysis. The task remains challenging because degraded documents contain uneven illumination, bleed-through, stains, faded ink, textured backgrounds, and weak foreground strokes. Classical methods such as Otsu~\cite{otsu1979threshold} and Sauvola~\cite{sauvola2000adaptive} estimate global or local thresholds from image statistics, while later methods incorporate document-specific priors such as background estimation and stroke preservation~\cite{gatos2006adaptive}. Recent CNN-, transformer-, generative-, and diffusion-based methods improve on them~\cite{ronneberger2015unet,souibgui2022docentr,cicchetti2024nafdpm}, yet preserving thin, broken, and low-contrast strokes under severe degradation remains difficult. Mamba-based state space models have shown strong potential for efficient long-range feature modeling~\cite{gu2023mamba}, and their linear-time propagation is attractive for high-resolution document images where both global degradation context and local stroke details matter. However, Mamba-based document binarization has received limited attention, with only preliminary studies~\cite{azfar2024amsdb}. Moreover, applying state-space propagation to document images reduces the amplitude contrast between weak foreground cues and the background, particularly for faint ink traces and fragmented characters. To address this, we propose DeepMine-Mamba, a Mamba-based framework with an Anti-Dilution Gate; the name reflects the idea of deeply mining stroke details that are otherwise diluted during state-space propagation. Instead of treating Mamba as a plug-in module, we first measure what propagation does to stroke-level feature statistics, and then design a gate that responds to the measured effect. The main contributions are:

\begin{itemize}
\itemsep0pt
    \item We characterize empirically how Mamba-based state-space propagation reshapes feature statistics in document images: it compresses the foreground-background amplitude contrast from $1.86$ to $1.03$, monotonically in the effective receptive field, which suggests the effect is not specific to state space models.

    \item We propose an Anti-Dilution Gate whose modulation is computed directly from the propagation-induced feature difference $|x-m|$ and confined to stroke regions by a learned stroke prior. We adopt the four-directional scanning scheme of VMamba~\cite{liu2024vmamba} without modification; the contribution lies in the analysis of propagation-induced dilution and in the gate that responds to it, not in the scanning mechanism.

    \item We evaluate DeepMine-Mamba on DIBCO/H-DIBCO under a strict leave-one-year-out protocol using no data outside the benchmark series, reporting protocol-controlled evidence separately from cross-paper comparison, with computational cost, component-wise ablations under a fixed objective, and a hyper-parameter sensitivity study.
\end{itemize}

\begin{figure*}[t]
\centering
\begin{adjustbox}{max width=0.98\textwidth, max height=0.26\textheight}    \input{fig_architecture}
\end{adjustbox}
\caption{
Overall architecture of DeepMine-Mamba, including ConvNeXt encoding, Sobel guidance, Mamba state modeling, and anti-dilution refinement.
}
\label{fig:architecture}
\end{figure*}

\section{Related Work}
\subsection{Document image binarization}
Classical binarization thresholds on global or local image statistics~\cite{otsu1979threshold,sauvola2000adaptive} or adds document-specific priors such as background estimation and post-processing~\cite{gatos2006adaptive}; these methods are interpretable but degrade under severe bleed-through, stains, or texture. Learning-based methods recast the task as dense prediction, from U-Net~\cite{ronneberger2015unet} and transformer-based DocEnTr~\cite{souibgui2022docentr} to diffusion-based NAF-DPM~\cite{cicchetti2024nafdpm}, but preserving thin degraded strokes remains challenging.

\subsection{State space models and Mamba}
Mamba introduces selective state space modeling with input-dependent parameters and linear-time sequence processing~\cite{gu2023mamba}. Because binarization requires both global degradation awareness and local stroke preservation, Mamba-style models are an attractive direction. However, applying state propagation to fine-structure-sensitive document images compresses the amplitude margin that separates strokes from the background: foreground responses are not attenuated in absolute terms, but background responses rise faster, an effect measured directly in Section~\ref{sec:dilution}.

\subsection{Dense prediction and document analysis}
State space models are adapted to vision by scanning two-dimensional feature maps as one-dimensional sequences. VMamba~\cite{liu2024vmamba} introduces a four-directional cross-scan that traverses the map in both horizontal and both vertical directions and fuses the four directional states, which has become the standard building block for Mamba-based dense prediction. In fine-structure-sensitive tasks, Mamba-based designs have been applied to image restoration~\cite{guo2024mambair} and medical image segmentation~\cite{ruan2024vmunet}, where thin structures and weak boundaries dominate the error. A recurring pattern is that propagation is paired with a local branch or convolutional stem, suggesting that it alone does not preserve fine local detail.

Within document analysis, state space models have been applied to visually-rich document understanding, where a segment-first bidirectional scan replaces self-attention over long documents~\cite{hu2025docmamba}; such models operate on text segments and perform semantic labeling rather than pixel-level prediction, whereas strokes are separated from the background by their local amplitude margin, and therefore propagation-induced contrast compression bears on binarization far more directly. For binarization specifically we are aware of only one preliminary study. AMSDB-VMamba~\cite{azfar2024amsdb} combines a VMamba encoder with a U-shaped decoder and injects Difference-of-Gaussians features through the skip connections; that hand-designed high-frequency path accounts for 7.06 of its 88.42 FM. Approaches addressing the same difficulty without state space models also compensate for lost local detail through external priors: GDB~\cite{yang2024gdb} learns gating values weakly guided by an a priori mask and edge map, and FourBi~\cite{quattrini2024fourbi} enlarges the receptive field through Fast Fourier Convolutions with a parallel local branch. Two gaps motivate this work. First, existing Mamba-based binarization treats propagation as a drop-in encoder without analyzing its effect on stroke-level feature statistics. Second, the detail lost during aggregation is compensated only indirectly, through hand-designed frequency or edge priors, not by a signal derived from the aggregation itself. We differ on both points: we measure the effect of propagation on foreground-background amplitude contrast, and we introduce a gate driven by the propagation-induced difference $|x-m|$, a quantity defined only when sequential state-space propagation is present.

\section{Proposed Method}
\subsection{Overview}
Figure~\ref{fig:architecture} illustrates the framework: a U-shaped encoder-decoder with a ConvNeXt-Tiny encoder~\cite{liu2022convnet} for multi-scale feature extraction. The encoder produces hierarchical features \(\{f_1,f_2,f_3,f_4\}\), progressively fused by a skip-connected decoder to obtain \(d_2\). In parallel, a Sobel edge prior~\cite{sobel1968isotropic} is extracted and later used for detail-aware prediction. Before propagation, \(d_2\) is enhanced by local contrast recovery and weak-text rescue, yielding feature \(x\) that enters the DeepMine-Mamba block, which performs stroke-aware Mamba propagation and Anti-Dilution refinement.

\subsection{Mamba-Based Feature Propagation}
Given a block input \(x \in \mathbb{R}^{B \times C \times H \times W}\), the propagation module captures long-range structural dependencies. Since document images contain large background regions, propagation raises background responses disproportionately and compresses the stroke-background amplitude margin (Section~\ref{sec:dilution}). To reduce this, we estimate a stroke prior map \(P_s \in [0,1]^{B \times 1 \times H \times W}\) and use it to modulate the feature.

\(P_s\) is predicted from \(x\) by a lightweight convolutional branch: a \(3\times3\) convolution halving the channel width, batch normalization and a GELU activation, followed by a \(1\times1\) convolution to a single channel and a sigmoid. The branch receives no direct supervision and is trained end-to-end through the binarization objective; \(P_s\) is therefore a learned saliency map rather than an externally imposed prior. Its measured selectivity of \(2.4\) (Section~\ref{sec:dilution}) indicates that stroke-aligned behavior emerges from the objective alone. The modulation is
\begin{equation}
\label{eq:modulation}
x_{\mathrm{fg}} = x \odot P_s + \lambda_{\mathrm{bg}} x \odot (1-P_s),
\end{equation}
where \(\odot\) is element-wise multiplication and \(\lambda_{\mathrm{bg}}=0.1\) retains a small amount of contextual background information.

\(x_{\mathrm{fg}}\) is then processed by four-direction Mamba propagation and the directional representations are adaptively fused by a pixel-directional attention module, a \(1\times1\) convolution producing four per-pixel weights normalized by a softmax, to obtain \(m\). The state-space core uses \(d_{\mathrm{model}}=96\), \(d_{\mathrm{state}}=16\) and \(\mathrm{dt\_rank}=6\). The DeepMine-Mamba block combines four-direction propagation, directional fusion, and the Anti-Dilution Gate; the Mamba-only variant in the ablation removes the gate but keeps the propagation.

\subsection{Anti-Dilution Gate}
Although four-direction propagation aggregates long-range context, the fused representation \(m\) exhibits markedly reduced foreground-background amplitude contrast (Section~\ref{sec:dilution}). This is harmful for binarization, where thin strokes and faint ink traces are distinguished from the background mainly by their local amplitude margin. We therefore introduce an Anti-Dilution Gate that restores the difference between the locally derived feature \(x\) and the propagated feature \(m\), starting from a dilution importance map \(I \in [0,1]^{B \times 1 \times H \times W}\):
\begin{equation}
\label{eq:importance}
I = \sigma \left( \phi_d \left( [x, |x - m|] \right) \right),
\end{equation}
where \(\phi_d(\cdot)\) is a lightweight convolutional detecto, a \(1\times1\) convolution from 192 to 48 channels with batch normalization and GELU, followed by a \(3\times3\) convolution to a single channel, \([\cdot,\cdot]\) the channel-wise concatenation, and \(\sigma(\cdot)\) the sigmoid. The term \(|x-m|\) describes the local response change caused by propagation, letting the gate identify where the stroke-background margin has been compressed. To avoid amplifying background noise, \(I\) is modulated by the stroke prior:
\begin{equation}
\label{eq:gate}
\hat{I} = \alpha_g P_s \odot I,
\end{equation}
where \(\alpha_g=0.7\) is the maximum gate strength, so that \(\hat{I}\) is bounded above by \(\alpha_g\). The refined feature interpolates between \(m\) and \(x\):
\begin{equation}
\label{eq:refine}
y = m + \hat{I} \odot (x - m).
\end{equation}
When \(\hat{I}\) approaches zero, the output stays close to \(m\); where \(\hat{I}\) is large, Eq.~\eqref{eq:refine} interpolates back towards \(x\), removing the background context that propagation injected there. Because the removed component originates from the background, the absolute amplitude of stroke responses decreases; what is restored is the discriminability between stroke and background, not the amplitude itself.

\subsection{Prediction Head}
The refined feature is upsampled and fused with the Sobel edge prior, then passed through four lightweight refinement modules at 64 channels --- structure refinement (two \(3\times3\) convolutions, residual), local variance suppression (\(9\times9\) average pooling fused by \(1\times1\) convolution), confidence gating (depthwise \(3\times3\), then \(1\times1\) convolution and a sigmoid), and weak-text refinement (a depthwise branch, weight \(\lambda_w=0.25\)) --- which are auxiliary rather than independent contributions. Three heads then estimate foreground probability \(P_{fg}\), background probability \(P_{bg}\), and an adaptive threshold map \(T\), each \(3\times3\) convolution followed by a \(1\times1\) projection. With \(P_{clean}=P_{fg}(1-P_{bg})\), the soft binarization output is
\begin{equation}
\label{eq:soft}
\hat{Y}=\sigma\left(\gamma(P_{clean}-T)\right),
\end{equation}
where \(\gamma=20.0\) controls the sharpness.

\subsection{Loss Function}
Let \(Y\) be the ground-truth binary map and \(\hat{Y}\in(0,1)\) the soft output. The main binarization loss is
\begin{equation}
\label{eq:lbin}
\mathcal{L}_{bin}
=
\mathcal{L}_{BCE}
+
\lambda_t \mathcal{L}_{T}
+
\lambda_f \mathcal{L}_{FM}
+
\lambda_p \mathcal{L}_{pFM},
\end{equation}
combining binary cross-entropy with a Tversky loss~\cite{salehi2017tversky} and two F-measure terms. With \(TP=\sum_i \hat{Y}_i Y_i\), \(FP=\sum_i \hat{Y}_i(1-Y_i)\), \(FN=\sum_i(1-\hat{Y}_i)Y_i\), \(\mathrm{Prec}=TP/(TP+FP)\), \(\mathrm{Rec}=TP/(TP+FN)\) and \(\mathcal{L}_{T}=1-(TP+1)/\allowbreak(TP+\alpha FP+\beta FN+1)\),
\begin{equation}
\mathcal{L}_{FM}=1-\frac{2\,\mathrm{Prec}\,\mathrm{Rec}}{\mathrm{Prec}+\mathrm{Rec}},\qquad
\mathcal{L}_{pFM}=1-\frac{2\,\mathrm{Prec}\,\mathrm{Rec}_s}{\mathrm{Prec}+\mathrm{Rec}_s},
\end{equation}
with \(\alpha=0.55\), \(\beta=0.45\), and \(\mathrm{Rec}_s=\sum_i \hat{Y}_i S_i / \sum_i S_i\) a skeleton recall on \(S=\mathrm{Skel}(Y)\); because \(S\) retains only stroke cores, \(\mathcal{L}_{pFM}\) strongly penalizes missing stroke centers and receives the largest weight. Section~\ref{sec:ablation} reports an ablation in which both F-measure terms are removed. The probability heads are supervised by binary cross-entropy (\(\mathcal{L}_{prob}\)), and three auxiliary regularizers are added: a boundary false-positive penalty \(\mathcal{L}_{bd}=\frac{1}{N}\sum_i \hat{Y}_i\,\partial Y_i\) over the one-pixel ring \(\partial Y\) around ground-truth strokes, which discourages halo artifacts that inflate DRD; a threshold regularizer \(\mathcal{L}_{th}=\frac{1}{N}\sum_i (T_i-0.5)^2\) keeping \(T\) near \(0.5\) and a skeleton supervision loss \(\mathcal{L}_{skel}\) on a \(1\times1\) convolution of the block output against a down-sampled \(\mathrm{Skel}(Y)\), used during training only. The overall objective is
\begin{equation}
\mathcal{L}
=
\mathcal{L}_{bin}
+
\lambda_{prob}\mathcal{L}_{prob}
+
\lambda_{bd}\mathcal{L}_{bd}
+
\lambda_{th}\mathcal{L}_{th}
+
\lambda_{skel}\mathcal{L}_{skel},
\end{equation}
with \(\lambda_t=0.5\), \(\lambda_f=0.5\), \(\lambda_p=1.0\), \(\lambda_{prob}=0.3\), \(\lambda_{bd}=0.1\), \(\lambda_{th}=0.01\), \(\lambda_{skel}=0.2\).

\section{Experiments}
\subsection{Experimental Setup}
Experiments use PyTorch on a single NVIDIA RTX 5080 (16GB), with an ImageNet-pretrained ConvNeXt-Tiny backbone~\cite{liu2022convnet} producing channels 96, 192, 384 and 768. We crop \(512\times512\) patches, use batch size 4 with 4-step gradient accumulation, and optimize with AdamW~\cite{loshchilov2019decoupled} for 200 epochs at a maximum learning rate of \(2\times10^{-4}\) under mixed precision. Inference uses sliding windows of crop 512 and stride 256, and we apply DIBCO-style augmentation simulating bleed-through, paper texture, ink blot, JPEG artifacts, and local blur.

\subsection{Datasets and Protocols}
\label{sec:protocols}
Experiments are conducted on the DIBCO and H-DIBCO series from 2009 to 2019~\cite{gatos2009dibco,pratikakis2010hdibco,pratikakis2011dibco,pratikakis2012hdibco,pratikakis2013dibco,ntirogiannis2014hdibco,pratikakis2016hdibco,pratikakis2017dibco,pratikakis2018hdibco,pratikakis2019dibco}. We use a leave-one-year-out protocol: for each target year its images are excluded from training and used only for testing, and the remaining years are used for training. This produces ten evaluation settings and assesses cross-year generalization. No data outside the DIBCO/H-DIBCO series is used for training. DIBCO 2019 is the hardest year: its Set B consists of ancient Greek papyrus fragments whose fiber structures locally resemble strokes~\cite{pratikakis2019dibco}, and no papyrus-like samples appear in the other years, so models must generalize to an unseen substrate. Tables~\ref{tab:overall_comparison} and~\ref{tab:dibco2019_comparison} annotate each method's protocol: \emph{free}, training-free; \emph{LOYO}, leave-one-year-out; \emph{LOO$+$}, leave-one-out with additional non-DIBCO data; \emph{held out}, the reported year excluded from training, and \emph{held out$+$} the same plus non-DIBCO data; \emph{n/r}, split not reported. Parentheses in Table~\ref{tab:overall_comparison} give the number of years averaged.

\subsection{Evaluation Metrics}
We use the standard DIBCO metrics --- F-measure (FM), pseudo F-measure (p-FM), PSNR, and distance reciprocal distortion (DRD)~\cite{pratikakis2017dibco,lu2004distance} --- all computed with the official DIBCO tool. p-FM emphasizes stroke preservation by computing recall on the skeletonized ground truth, and DRD measures perceptual distortion from local errors.

\begin{table}[t]
\begin{center}
\caption{Leave-one-year-out results of DeepMine-Mamba on each DIBCO/H-DIBCO benchmark year, computed with the official DIBCO evaluation metrics.}
\label{tab:loo_yearly_results}
\resizebox{0.80\linewidth}{!}{
\begin{tabular}{lcccc}
\toprule
Test Year & FM$\uparrow$ & p-FM$\uparrow$ & PSNR$\uparrow$ & DRD$\downarrow$\\
\midrule
2009 & 95.74 & 96.50 & 21.59 & 1.32 \\
2010 & 94.47 & 96.84 & 21.83 & 1.35 \\
2011 & 95.35 & 97.56 & 21.40 & 1.36 \\
2012 & 96.26 & 97.32 & 23.30 & 1.19 \\
2013 & 96.15 & 97.93 & 23.13 & 1.23 \\
2014 & 97.45 & 98.84 & 23.58 & 0.69 \\
2016 & 90.28 & 94.90 & 19.26 & 3.25 \\
2017 & 93.31 & 95.19 & 19.39 & 2.35 \\
2018 & 90.69 & 94.62 & 20.00 & 2.87 \\
2019 & 75.90 & 76.44 & 15.41 & 7.82 \\
\midrule
Average & 92.56 & 94.61 & 20.88 & 2.34 \\
\bottomrule
\end{tabular}
}
\end{center}
\end{table}

\subsection{Comparison with Existing Methods}
\label{sec:comparison}
Published results on the DIBCO series are not obtained under a common protocol: training corpora, year coverage, and whether a given year is held out all vary between papers, and for most entries in Table~\ref{tab:dibco2019_comparison} the split is not stated. FourBi and GDB do document their settings, but both train outside the DIBCO series, and therefore re-running them under a DIBCO-only protocol would not evaluate the published models; the main comparison in Table~\ref{tab:overall_comparison} is nonetheless protocol-controlled, being taken from a common leave-one-year-out re-evaluation~\cite{biswas2023docbinformer}. Averages over different numbers of years are likewise not interchangeable: omitting DIBCO 2019 raises our own average FM from 92.56 to 94.41. We therefore report all of our own results under a single leave-one-year-out protocol, annotate each baseline's protocol (Section~\ref{sec:protocols}), and treat the cross-paper tables as benchmark-level references; Tables~\ref{tab:loo_yearly_results},~\ref{tab:ablation_adg}, and~\ref{tab:loss_hyper} give the protocol-controlled evidence.

\begin{table}[t]
\centering
\caption{
Benchmark-level comparison; protocol labels follow Section~\ref{sec:protocols}, and dashes mark training-free methods or values the source does not report. Baselines in the main comparison other than GDB come from the leave-one-year-out re-evaluation in~\cite{biswas2023docbinformer} rather than their original papers, GDB from its own paper; DocEnTr covers nine of the ten years, and methods absent from that re-evaluation (DocDiff, NAF-DPM, DE-GAN) appear only in Table~\ref{tab:dibco2019_comparison}. Methods with different year coverage are listed separately, each followed by our score on the same years. GDB additionally trains on Bickley Diary, PHIBD and S-MS, FourBi on Bickley Diary, SMADI and Nabuco; ours are on nothing outside DIBCO/H-DIBCO.
}
\label{tab:overall_comparison}
\resizebox{\linewidth}{!}{
\begin{tabular}{lcccccc}
\toprule
Method & Params & FM$\uparrow$ & p-FM$\uparrow$ & PSNR$\uparrow$ & DRD$\downarrow$ & Protocol \\
\midrule
Otsu~\cite{otsu1979threshold} & -- & 77.27 & 79.14 & 15.22 & 17.24 & free (10) \\
Sauvola~\cite{sauvola2000adaptive} & -- & 78.10 & 82.91 & 15.77 & 8.72 & free (10) \\
GiB~\cite{bhowmik2019gib} & -- & 87.33 & 89.85 & 17.98 & 5.40 & free (10) \\
RDD~\cite{su2013robust} & -- & 85.96 & 87.43 & 17.92 & 7.00 & free (10) \\
DeepOtsu~\cite{he2019deepotsu} & -- & 88.59 & 90.84 & 19.33 & 3.76 & LOYO (10) \\
cGANs~\cite{zhao2019cascaded} & -- & 90.24 & 91.97 & 19.66 & 3.81 & LOYO (10) \\
DocEnTr~\cite{souibgui2022docentr} & 68.8\,M & 90.51 & 92.17 & 20.31 & 1.15 & LOYO (9) \\
DocBinFormer~\cite{biswas2023docbinformer} & -- & 92.22 & 94.09 & 21.30 & \textbf{0.15} & LOYO (10) \\
D$^2$BFormer~\cite{yang2023d2bformer} & -- & 91.75 & 93.05 & 20.85 & 2.71 & LOYO (10) \\
TransDocUNet~\cite{sukesh2023transdocunet} & 105\,M & 92.44 & 93.86 & \textbf{22.16} & 0.49 & LOYO (10) \\
GDB~\cite{yang2024gdb} & -- & 92.04 & 94.01 & 20.50 & 2.85 & LOO$+$ (10) \\
DeepMine-Mamba & 33.3\,M & \textbf{92.56} & \textbf{94.61} & 20.88 & 2.34 & LOYO (10) \\
\midrule
\multicolumn{7}{@{}l}{\emph{Different year coverage --- matched-year comparison}}\\
FourBi~\cite{quattrini2024fourbi} & 17.6\,M & 95.34 & 98.14 & 22.45 & 1.50 & LOO$+$ (8) \\
\quad ours, same 8 years & & 94.25 & 96.65 & 21.49 & 1.79 & LOYO (8) \\
AMSDB-VMamba~\cite{azfar2024amsdb} & 59\,M & 88.42 & 93.10 & 18.27 & -- & n/r (1) \\
\quad ours, DIBCO 2016 & & 90.28 & 94.90 & 19.26 & 3.25 & LOYO (1) \\
\bottomrule
\end{tabular}
}
\end{table}

As shown in Table~\ref{tab:overall_comparison}, DeepMine-Mamba obtains 92.56 average FM and 94.61 p-FM, above the other ten-year entries, and improves on GDB in all four metrics. The FM margins are narrow: 0.12 over TransDocUNet does not exceed the retraining shift of Section~\ref{sec:ablation}, and 0.34 over DocBinFormer only just does. The p-FM margins are larger, 0.75 over TransDocUNet and 0.52 over DocBinFormer, but retraining variation was measured for FM only, and therefore we do not read them as separations either. It does not lead on PSNR or DRD, where TransDocUNet reports 22.16 and DocBinFormer 0.15, and the two entries reporting DRD below $0.6$ here are the same ones whose DRD is shown below to be inconsistent. Since the two metrics on which it does lead are moreover aligned with two loss terms in Eq.~\eqref{eq:lbin}, we treat Tables~\ref{tab:overall_comparison} and~\ref{tab:dibco2019_comparison} as benchmark-level references rather than rankings, remove those terms explicitly in Section~\ref{sec:ablation}, and rest the DRD and PSNR claims of this paper on Table~\ref{tab:ablation_adg}, where the gate lowers DRD from 4.79 to 2.59 and raises PSNR from 19.72 to 20.16 under a fixed objective and protocol. Table~\ref{tab:overall_comparison} also includes two methods with different year coverage. FourBi~\cite{quattrini2024fourbi} covers 2010--2018 under leave-one-out with three external corpora, and over exactly those eight years it is ahead of DeepMine-Mamba on all four metrics. On DIBCO 2019, which it excludes from training entirely, the ordering reverses (Table~\ref{tab:dibco2019_comparison}), despite FourBi having strictly more training data available there than we do. AMSDB-VMamba~\cite{azfar2024amsdb}, the only prior Mamba-based binarization model, reports DIBCO 2016 only and trains on 2009--2014, whereas our setting uses all nine remaining years; there DeepMine-Mamba leads on all three metrics it reports, at 33.3\,M parameters against 59\,M. Published runtimes use different hardware and input sizes and are not directly comparable: 190 ms and 90 ms per image for TransDocUNet and AMSDB-VMamba, 3.8 ms and 9.1 ms at $256^2$ for DocEnTr and FourBi, against 20.3 ms per $512\times512$ crop for ours.

\begin{table}[t]
\centering
\caption{
Comparison on DIBCO 2019, restricted to methods publishing in this year (GiB, RDD and cGANs do not). Protocol labels follow Section~\ref{sec:protocols}; numbers are from the respective publications, except DE-GAN ($\dagger$), taken from the re-evaluation in~\cite{quattrini2024fourbi} since its own paper does not report this year. Entries labeled LOYO in Table~\ref{tab:overall_comparison} appear as n/r because that re-evaluation reports series averages only, and therefore their per-year scores here come from their own papers, which do not state the split. DocDiff and NAF-DPM do not report DRD, which is not highlighted because the reported values are not mutually consistent (see text).
}
\label{tab:dibco2019_comparison}
\resizebox{\linewidth}{!}{
\begin{tabular}{lccccl}
\toprule
Method & FM$\uparrow$ & p-FM$\uparrow$ & PSNR$\uparrow$ & DRD$\downarrow$ & Protocol \\
\midrule
Otsu~\cite{otsu1979threshold}            & 63.87 & 60.24 & 12.67 & 16.87 & free \\
Sauvola~\cite{sauvola2000adaptive}       & 63.82 & 60.18 & 12.66 & 16.91 & free \\
DeepOtsu~\cite{he2019deepotsu}           & 60.75 & 59.91 & 14.44 & 13.46 & n/r \\
DE-GAN$^\dagger$~\cite{souibgui2020degan}         & 57.96 & 57.30 & 14.43 & 12.21 & n/r \\
DocEnTr~\cite{souibgui2022docentr}       & 59.00 & 60.00 & 13.85 & 0.30  & n/r \\
DocDiff~\cite{yang2023docdiff}           & 73.38 & 75.12 & 15.14 & --    & n/r \\
DocBinFormer~\cite{biswas2023docbinformer} & 60.31 & 64.00 & 14.49 & 0.21 & n/r \\
TransDocUNet~\cite{sukesh2023transdocunet} & 56.79 & 56.78 & 13.36 & 0.56 & n/r \\
D$^2$BFormer~\cite{yang2023d2bformer}    & 67.63 & 66.69 & 15.05 & 10.59 & n/r \\
NAF-DPM~\cite{cicchetti2024nafdpm}       & 74.61 & 76.25 & 15.39 & --    & n/r \\
GDB~\cite{yang2024gdb}                   & 73.51 & 74.81 & 14.96 & 10.75 & LOO$+$ \\
FourBi~\cite{quattrini2024fourbi}        & 63.56 & \textbf{85.82} & 15.02 & 9.66 & held out$+$ \\
\midrule
DeepMine-Mamba  & \textbf{75.90} & 76.44 & \textbf{15.41} & 7.82 & LOYO \\
\bottomrule
\end{tabular}
}
\end{table}

On DIBCO 2019 (Table~\ref{tab:dibco2019_comparison}), three entries document that the year was excluded from training, DeepMine-Mamba, GDB and FourBi, while the rest do not state the split and should be read with that caveat. Among the three, DeepMine-Mamba leads on FM at 75.90 against 73.51 and 63.56, and also on PSNR and DRD, while FourBi leads on p-FM by 9.38; it is moreover the only one of the three trained without data outside the benchmark series (Table~\ref{tab:overall_comparison}). FourBi's high p-FM with a much lower FM indicates predictions covering stroke cores but under-covering stroke bodies, since p-FM rewards core recall more than FM does. Across all entries listed, DeepMine-Mamba attains the largest FM; its PSNR of 15.41 is nominally the largest but lies within 0.02 of NAF-DPM and should not be read as a separation. Reported DRD values fall into two disjoint groups: three transformer-based methods report values below $0.6$ while scoring below 61 FM, whereas every method above 63 FM that reports DRD, including GDB, FourBi, and ours, gives a value between $7.8$ and $17$. A low FM does not by itself produce a low DRD: DeepOtsu and DE-GAN also score below 61 FM yet report 13.46 and 12.21. An FM 20 points lower with a twenty-fold lower DRD is hard to reconcile within one evaluation procedure. FourBi's re-evaluation of DocEnTr on this year supports that reading, moving it from 59.00 FM, 60.00 p-FM, and 0.30 DRD to 67.20, 77.13, and 10.20~\cite{quattrini2024fourbi}. We therefore compare DRD only among mutually consistent entries; there DeepMine-Mamba records the lowest value, 7.82 against FourBi's 9.66. All our scores use the official DIBCO evaluation tool.

\begin{table}[t]
\centering
\caption{Component-wise leave-one-year-out ablation under an identical objective and protocol. The fourth row is the controlled state-space baseline, differing from the fifth only by the absence of the gate.}
\label{tab:ablation_adg}
\resizebox{\linewidth}{!}{
\begin{tabular}{lcccc}
\toprule
Model & FM$\uparrow$ & p-FM$\uparrow$ & PSNR$\uparrow$ & DRD$\downarrow$\\
\midrule
Baseline U-Net & 74.84 & 75.55 & 16.74 & 8.13 \\
+ ConvNeXt Encoder & 86.17 & 87.96 & 17.39 & 7.24 \\
+ Sobel Edge Prior & 87.05 & 88.53 & 17.78 & 6.51 \\
+ 4-dir Mamba, VMamba-style & 90.54 & 93.35 & 19.72 & 4.79\\
+ Anti-Dilution Gate & 92.43 & 94.27 & 20.16 & 2.59 \\
+ Refinement Heads & \textbf{92.56} & \textbf{94.61} & \textbf{20.88} & \textbf{2.34}\\
\bottomrule
\end{tabular}
}
\end{table}

\section{Ablation and Analysis}
\subsection{Ablation Study}
\label{sec:ablation}
In Table~\ref{tab:ablation_adg}, the ConvNeXt encoder and Sobel edge prior together raise FM to 87.05 and reduce DRD to 6.51, and four-direction Mamba propagation improves FM to 90.54, confirming the benefit of long-range state-space modeling. Adding the Anti-Dilution Gate raises FM to 92.43 and reduces DRD from 4.79 to 2.59; the refinement heads give the final 92.56 FM, 94.61 p-FM, 20.88 PSNR, and 2.34 DRD. The gate improves PSNR and DRD as well as the two F-measure metrics, DRD falls by 46\% and PSNR rises by 0.44, and therefore its benefit is not confined to the quantities that appear in the objective.

\begin{table}[t]
\centering
\caption{Gate contribution with and without the F-measure-aligned loss terms, under the same ten-year leave-one-year-out protocol as Table~\ref{tab:ablation_adg}; all four variants are retrained, and therefore the with-terms column repeats the fourth and fifth rows of that table to within 0.25 FM. Only FM is reported, since the removed terms are F-measure surrogates.}
\label{tab:loss_hyper}
\resizebox{\linewidth}{!}{
\begin{tabular}{lccc}
\toprule
FM & with gate & without gate & gain \\
\midrule
with $\mathcal{L}_{FM}$, $\mathcal{L}_{pFM}$ & 92.2 & 90.7 & \textbf{+1.5} \\
without & 91.5 & 90.4 & \textbf{+1.1} \\
\bottomrule
\end{tabular}
}
\end{table}

All variants in Table~\ref{tab:ablation_adg} are trained with an identical objective and protocol, and therefore \(\mathcal{L}_{FM}\) and \(\mathcal{L}_{pFM}\) are held constant across every row and the improvement attributable to the gate cannot arise from the F-measure-aligned loss terms. These two terms are moreover only differentiable surrogates: they do not reproduce the official DIBCO definitions, whose pseudo-recall and pseudo-precision are weighted. To test the dependence directly, Table~\ref{tab:loss_hyper} removes both terms and retrains. The gate contribution in the last column falls from \(+1.5\) to \(+1.1\) FM between the two rows but does not disappear, so its benefit does not depend on the alignment; within the with-gate column, removing both terms costs the gated variant 0.7 FM, which bounds the extent to which the absolute scores of Table~\ref{tab:overall_comparison} rest on that alignment. Repeating a variant under the same protocol shifted FM by at most 0.25 in the two cases we measured, namely the with-terms column of Table~\ref{tab:loss_hyper} against rows four and five of Table~\ref{tab:ablation_adg}; the gate's \(+1.5\) to \(+1.9\) clears that shift while the \(+0.13\) from the refinement heads does not, which is why we call them auxiliary.

\noindent\textbf{Hyper-parameter sensitivity} We restrict the study to the two hyper-parameters introduced by the gate; the loss weights, Tversky coefficients and backbone configuration follow common practice and were not tuned per fold. On a three-fold subset chosen as the three hardest years of the series (2016, 2018, 2019), again with leave-one-year-out within each fold, \(\alpha_g\) yields 84.69, 84.83, 85.19, and 84.67 FM at \(0.3\), \(0.5\), \(0.7\), and \(0.9\); performance is stable and we adopt \(\alpha_g=0.7\). The small decline at \(0.9\) is in the direction Eq.~\eqref{eq:refine} predicts: raising \(\alpha_g\) raises the upper bound on \(\hat{I}\) and therefore discards more of the propagated representation at stroke locations. Retaining a small amount of background context before propagation is necessary: \(\lambda_{\mathrm{bg}}=0\), \(0.05\), \(0.1\), and \(0.2\) give 83.82, 85.02, 85.19, and 84.72 FM, and therefore removing it entirely costs 1.37 FM, while performance varies by less than 0.5 FM between \(0.05\) and \(0.2\). These are separate training runs over three years, one of them out-of-distribution, and therefore run-to-run variation here exceeds the 0.25 measured on ten-year averages; absolute values are correspondingly not comparable with Table~\ref{tab:loo_yearly_results} or~\ref{tab:ablation_adg}, and the \(\alpha_g\) range is read as stability rather than as a preference for \(0.7\). The effect concentrates on the out-of-distribution year: over \(\lambda_{\mathrm{bg}}\), FM varies by at most 1.1 on 2016 and 2018 but by 5.2 on 2019, where fiber structures resemble strokes and background context acts as a corrective signal.

\begin{figure*}[t]
\centering
\includegraphics[width=0.88\textwidth]{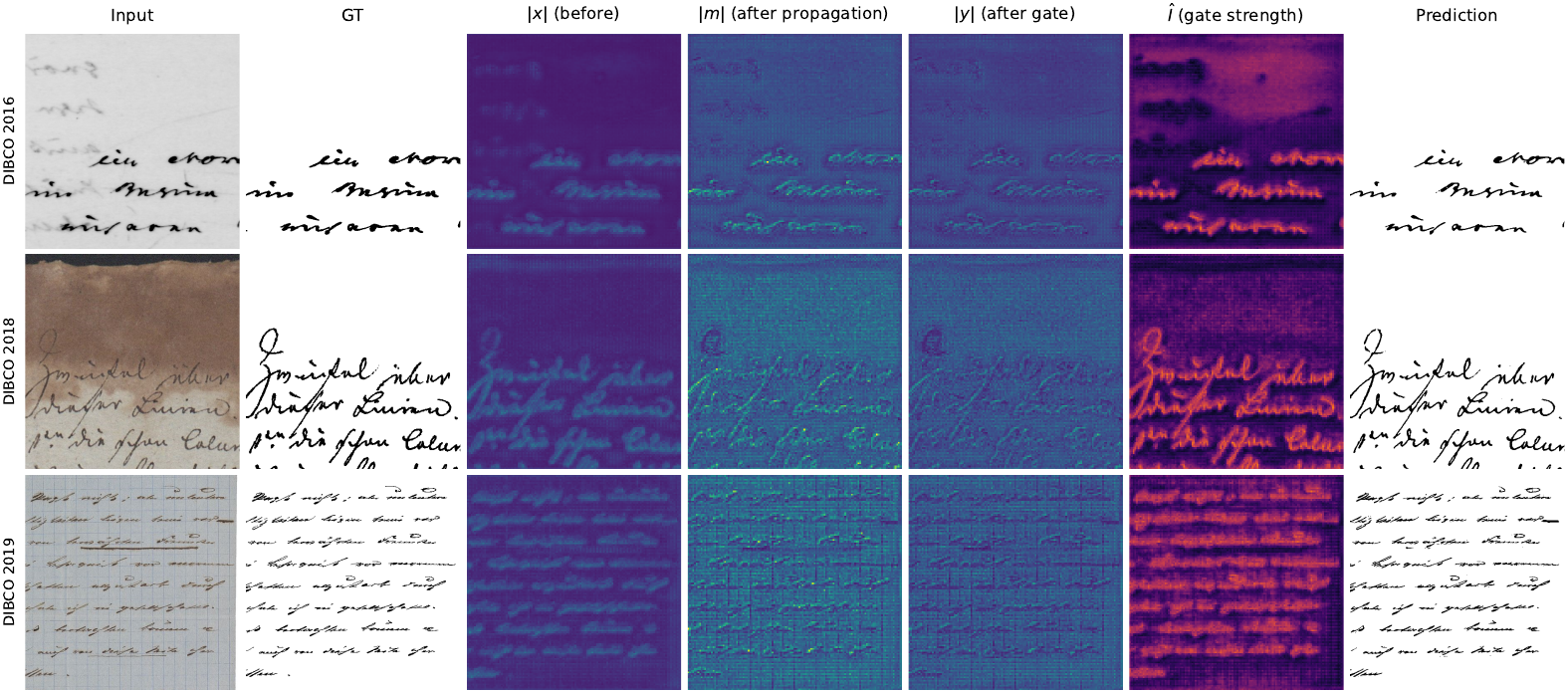}
\caption{
Feature maps at three stages of the DeepMine-Mamba block on DIBCO 2016, 2018 and 2019. Columns: input, ground truth, \(|x|\) before propagation, \(|m|\) after propagation, \(|y|\) after the gate, gate strength \(\hat{I}\), and prediction. Stroke regions stand out clearly before propagation and become indistinguishable after it as background responses rise to their level; the gate re-concentrates on strokes (mean \(\hat{I}\) \(0.327\) against \(0.137\) in the background) and the prediction column shows faint strokes on papyrus recovered while fiber-rich background is suppressed. On DIBCO 2019 (bottom row) \(\hat{I}\) also responds to substrate fibers, the limitation of Section~\ref{sec:limitations}.
}
\label{fig:dilution_maps}
\end{figure*}

\subsection{Characterizing Propagation-Induced Dilution}
\label{sec:dilution}

\noindent\textbf{Setup} We analyze the trained models on DIBCO 2016, 2018 and 2019 under the leave-one-year-out protocol, extracting four \(512\times512\) crops per image. Features are taken at three points of the block, which runs at stride 4 with 96 channels: \(x\) entering it, \(m\) after directional fusion, and \(y\) after the gate. The ground truth is mapped to this resolution by \(4\times4\) pooling; a cell is a stroke cell if it contains any foreground pixel and a background cell if it contains none and is not adjacent to one under \(3\times3\) dilation, and therefore boundary cells contaminate neither group. The amplitude \(a(p)\) is the \(L_2\) norm across the 96 channels.

\noindent\textbf{Contrast dilution} Propagation does not attenuate foreground responses in absolute terms, every region gains amplitude. The gain is strongly unequal, and the ratio of the two gains is invariant to any global rescaling of the feature map, since \(\mathrm{gain}_{bg}/\mathrm{gain}_{fg} = C(x)/C(m)\) with \(C = a_{fg}/a_{bg}\). The background gain is \(1.8\) times the stroke gain, and \(C\) falls from \(1.86\) to \(1.03\), i.e.\ to within 3\% of the value at which the two classes are indistinguishable in amplitude (Table~\ref{tab:dilution}). This is dilution in the literal sense: the discriminative signal is not removed but diluted by a disproportionate rise in the background response. Only the within-map ratio \(C\) is scale-invariant, since \(x\) and \(m\) come from different operations.

\noindent\textbf{Where the information goes} A linear probe on the full 96-dimensional feature achieves an AUC above \(0.99\) at all three stages, confirming that propagation does not remove foreground information. What it removes is the amplitude margin that makes stroke regions locally salient. Because a downstream readout can learn either polarity, the usable amplitude cue is the deviation of the ROC area from chance: propagation reduces \(|\mathrm{AUC}-0.5|\) from \(0.470\) to \(0.035\), retaining only 7\% of the original amplitude discriminability, and the gate recovers it to \(0.152\), about one third of the original. After the gate, stroke cells carry lower amplitude than background cells: the polarity inverts while discriminability increases, which is immaterial for the prediction heads. This follows from Eq.~\eqref{eq:refine}. Propagation raises every amplitude, and therefore mixing \(x\) back into \(m\) lowers it, and the gate does so far more at stroke locations, mean \(\hat{I}\) of \(0.327\) there against \(0.137\) in the background, a selectivity of \(2.4\) confirming that Eq.~\eqref{eq:gate} confines compensation to strokes, which drives \(C\) below 1 while separating the two groups further. Figures~\ref{fig:dilution_maps} and~\ref{fig:dilution_dist} show the corresponding maps and amplitude distributions.

\noindent\textbf{Is this confined to state space models?} Replacing the propagation step with fixed mean filters of increasing kernel size and recomputing \(C\) shows that the retained contrast margin \((C-1)/(C(x)-1)\), evaluated before rounding, decreases monotonically with the receptive field, from 68.8\% at \(5\times5\) to 0.3\% at \(63\times63\), with four-directional propagation at 3.1\%, closest to a \(31\times31\) kernel at feature resolution. Mean filters are untrained, and therefore this establishes that contrast dilution scales with the aggregation radius, instead which the two are interchangeable. Attention-based aggregation is left to future work: unlike a mean filter it cannot be substituted into an already-extracted feature map, and therefore the comparison would require training separate models. Within this scope, contrast dilution appears to be a general consequence of large-receptive-field aggregation rather than a defect of state-space propagation, and therefore the compensation should apply beyond Mamba-based architectures. This does not imply that a smaller receptive field would be preferable: Table~\ref{tab:ablation_adg} shows that four-directional propagation improves FM from 87.05 to 90.54, at a cost of 0.91 G of the 115.70 G multiply-accumulate operations at $512\times512$, 0.8\%, the selective scan being uncounted by standard profilers and added analytically as $9LDN$ per direction, and 0.068 M parameters, 0.2\% of a model dominated by the ConvNeXt encoder (27.82 M of 33.32 M). Full-image latency is a median of 324 ms [IQR 153--780], or 274 ms/MP after area normalization, with 704 MB peak memory; the sliding window of crop 512 and stride 256 evaluates every pixel four times, as FourBi's does~\cite{quattrini2024fourbi}, and therefore this cost belongs to the inference protocol rather than to state-space propagation.

\begin{table}[t]
\centering
\caption{Foreground-background amplitude statistics before propagation (\(x\)), after propagation (\(m\)), and after the gate (\(y\)). Separability is deviation from chance, since a downstream readout can learn either polarity; after the gate the polarity inverts while discriminability increases. \(|d|\) is the absolute standardized mean difference between the groups.}
\label{tab:dilution}
\resizebox{\linewidth}{!}{
\begin{tabular}{lccc}
\toprule
Measure & \(x\) & \(m\) & \(y\) \\
\midrule
Amplitude contrast \(C=a_{fg}/a_{bg}\) & 1.86 & 1.03 & 0.89 \\
Amplitude separability \(|\mathrm{AUC}-0.5|\) & 0.470 & 0.035 & 0.152 \\
Amplitude separability \(|d|\) & 2.72 & 0.07 & 0.40 \\
Linear-probe AUC (96-D) & 0.991 & 0.993 & 0.994 \\
\bottomrule
\end{tabular}
}
\end{table}

\subsection{Limitations and Future Work}
\label{sec:limitations}
The main limitation follows from the stroke prior. Because \(P_s\) receives no direct supervision and appears in both Eq.~\eqref{eq:modulation} and Eq.~\eqref{eq:gate}, the gate amplifies precisely those locations the model already believes to be foreground. On in-distribution years this is benign, but DIBCO 2019 Set B contains papyrus fibers whose local appearance resembles strokes; \(P_s\) responds to them, and Eq.~\eqref{eq:gate} grants them a high gate strength; the restored local responses are then read out as isolated foreground, which is consistent with the elevated DRD of 7.82 that year and visible in the bottom row of Figure~\ref{fig:dilution_maps}. The effect is confined to that year: averaged over the nine remaining benchmark years, FM is 94.41, DRD 1.73, and PSNR 21.50, against 92.56, 2.34, and 20.88 over all ten. These figures locate the source of the gap rather than establishing a comparison, since the baselines are not re-averaged the same way; the sensitivity results agree \(\lambda_{\mathrm{bg}}\) affecting 2019 five times more than the in-distribution years. Supervising \(P_s\) directly, or conditioning the gate on a substrate estimate, is therefore a natural next step. The gate should also be evaluated on other dense prediction tasks where large-receptive-field aggregation compresses local contrast.

\noindent
\begin{minipage}{\linewidth}
\vspace{2pt}
\centering
\includegraphics[width=\linewidth]{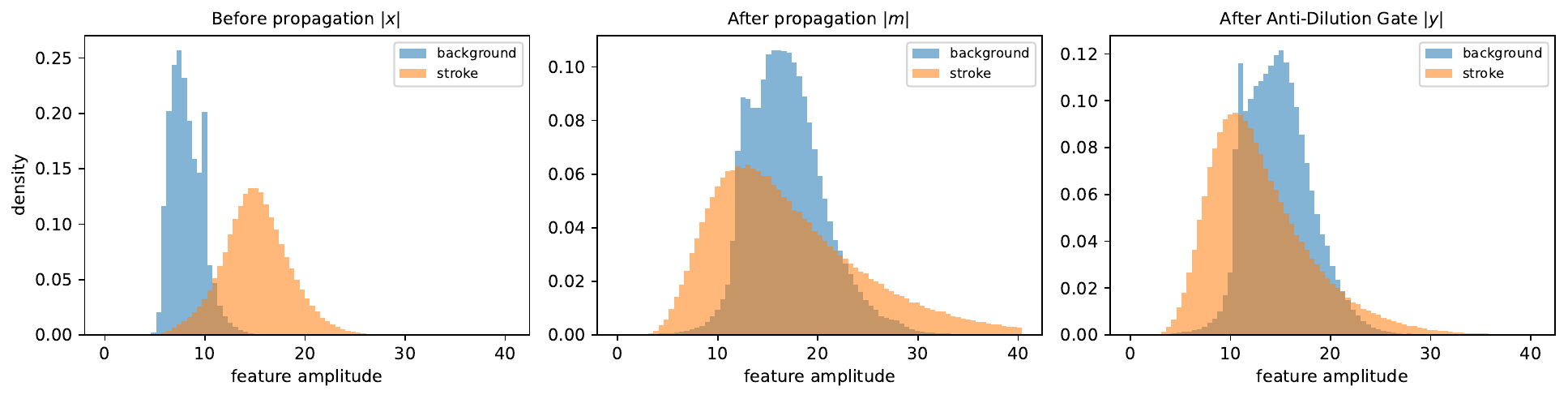}
\captionof{figure}{
Amplitude distributions of stroke and background cells before propagation, after propagation, and after the Anti-Dilution Gate.
}
\label{fig:dilution_dist}
\end{minipage}

\section{Conclusion}
We present DeepMine-Mamba, a Mamba-based binarization framework. State-space propagation does not attenuate foreground responses but compresses the foreground-background amplitude contrast from \(1.86\) to \(1.03\) through a disproportionate rise in background response, and this compression scales with the effective receptive field, and therefore it is unlikely to be specific to state space models. The Anti-Dilution Gate derives its modulation from the propagation-induced feature difference and confines it to stroke regions, recovering about one third of the original discriminability. Under a strict leave-one-year-out protocol with no external data it is competitive on DIBCO/H-DIBCO, and ablations under a fixed objective isolate the gate as the source of the improvement on all four metrics. Compensating for aggregation-induced dilution thus looks promising for adapting state space models to fine-structure-sensitive dense prediction.

\section*{Acknowledgment}
\noindent This research work is partially supported by the National Science and Technology Council (NSTC), Taiwan, under grant number 114-2221-E-032-011.

\vspace{3pt}
\noindent\textbf{Declaration of generative AI use.} The authors used a large language model for language editing and clarity refinement; all content was reviewed by the authors, who take full responsibility.

\setlength{\bibsep}{-0.3pt}
\bibliographystyle{model1-num-names}
\bibliography{references}

\end{document}